# Spatiotemporal KSVD Dictionary Learning for Online Multi-target Tracking

Huynh Manh
Department of Computer Science and Engineering
University of Colorado Denver
Denver, Colorado, USA
huynh.manh@ucdenver.edu

Gita Alaghband
Department of Computer Science and Engineering
University of Colorado Denver
Denver, Colorado, USA
gita.alaghband@ucdenver.edu

*Abstract*— In this paper, we present a new spatiotemporal discriminative KSVD dictionary algorithm (STKSVD) for learning target appearance in online multi-target tracking system. Different from other classification/recognition tasks (e.g. face, image recognition), learning target's appearance in online multi-target tracking is impacted by factors such as: posture/articulation changes, partial occlusion by background scene or other targets, background changes (human detection bounding box covers both human parts and part of the scene), etc. However, we observe that these variations occur gradually relative to spatial and temporal dynamics. We characterize the spatial and temporal information between target's samples through a new STKSVD appearance learning algorithm to better discriminate targets. Our STKSVD method is able to learn discriminative sparse code, linear classifier parameters, and minimize reconstruction error in single optimization system. Our appearance learning algorithm and tracking framework employs two different methods of calculating appearance similarity score in each stage of a two-stage association: a linear classifier in the first stage, and minimum residual errors in the second stage. The results tested using 2DMOT2015 dataset and its public Aggregated Channel Features (ACF) human detection for all comparisons show that our method outperforms the existing related learning methods.

## I. INTRODUCTION

Multi-target tracking is one of the computer vision challenging problems. The task is to keep track of all targets, humans in our case, in a scene (i.e., a video sequence) and maintain their identities throughout their presence in the sequence. Multi-target tracking systems have applications in robot navigation [1], surveillance systems [2],video analysis [2] and autonomous driving cars [3]. Despite great improvements achieved in tracking performance, there remains challenges arising from factors such as: (1) Occlusion, i.e., people often partially or fully occlude each other while moving or are occluded by background and other objects (e.g., trees, cars, etc.). (2) Motion prediction, i.e., unpredictable movement of targets involving linear and non-linear motion which can also cause large variations in appearance. (3) Discriminative appearance, i.e., distinguishing individuals with similar appearance (e.g., same clothes' color). Of course, other factors like abrupt camera motions, light changes, etc. will also impact tracking systems. Since it is hard to solve all of the above problems in one unified tracking framework, researchers often focus on either motion modeling [4]–[6], appearance learning [7]–[9] or tracking systems that combine a subset of tracking components (i.e., motion, shape, appearance) [7], [10].

An efficient target appearance learning model plays a very important role in tracking systems. Several appearance learning methods have been used in variety of tracking systems such as incremental linear discriminant analysis (ILDA) [7], incremental/decremental support vector machine (ID-SV) [11], label consistent KSVD (LCKSVD) [8], and reinforcement learning [10]. These methods apply well-known learning algorithms in multi-target tracking systems without considering the spatial and temporal relationship between target's samples. Most existing tracking systems calculate motion, location, and appearance similarity separately to form a final similar score, and then apply a data association (e.g., Hungarian algorithm [12] or greedy algorithm [11]) to find matching pairs between detections and targets. We note that the spatial and temporal information must be inherently encoded in appearance changes. We also observe several differences between appearance learning in other image recognition/classification tasks (e.g., face recognition) and in multi target tracking. For example, face images are usually fixed with frontal faces while target appearances have much larger variances along with nonlinear motions. In addition, human detection bounding box (encompassing rectangle generated by human detectors) does not only cover target's appearance but also parts of the background scene, making the task of appearance modeling more difficult.

In this work, we propose a novel online appearance learning algorithm, based on discriminative KSVD dictionary learning, that incorporates the changes with respect to target's location and time, called spatiotemporal KSVD algorithm (STKSVD). The reason for choosing the KSVD dictionary learning method as the baseline for our appearance learning are two-fold: (1) the original KSVD algorithm [13] and its improved versions of discriminative learning [14], [15] have been successful in handling partial occlusions in image recognition. (2) KSVD algorithm or other dictionary learning algorithms, in general, operate on dictionary columns (i.e., atoms), which will be learned/trained through a learning step. One can imagine that one atom represents one target's sample at a specific time. Thus, it is a suitable base model for our purpose to characterize and encode the spatial and temporal relationship between these different atoms and between atoms and training data. Thereby providing the ability to exploit the locality information pertinent to each target at a given time and location. Our method aims to learn spatially and temporally discriminative sparse code and a set of linear classifier parameters to further reduce classification error and signal reconstruction error. We also combine our learning appearance with location similarity and shape similarity for calculating the final similarity score. In summary, our contributions are:

- A new spatiotemporal KSVD algorithm (STKSVD) for learning online target appearance in multi-target tracking. The model encodes the spatial and temporal relationship between training data and dictionary atoms making sparse representation of human detections more discriminative.
- Implementing two different scenarios for calculating appearance similarity scores for each of the two association stages. It is done by passing the spare code of a detection into a linear classifier in the first stage and calculating the minimum residual error in the second stage of association.

The remaining of the paper is organized as follows. In section II, we review the related work. We will discuss our system design in section III and present the new spatiotemporal STKSVD dictionary learning in section IV. We present our results in section V followed by a discussion of our system limitations and future work in section VI.

## II. RELATED WORKS

**Multi-target Tracking by Detection**: One can classify multi-target tracking into different categories depending on the characteristics of tracking systems, namely tracking by detections [10] versus tracking without detections [16] combined with online tracking [7] versus batch tracking [17] which can be further combined with or without offline motion/appearance learning [10]. Based on this information, we classify our tracking system into online tracking-by-detections, without offline learning. Tracking-by-detection framework uses human detection sets as its input. Thus, tracking performances are highly dependent on the accuracy of the human detector algorithm employed. Although recent tracking-by-detection systems have shown to be superior to tracking-by-non-detection methods, their accuracy are still low overall. In this work, we opt to use public Aggregate Channel Features (ACF) detector [18] that has produced the most reliable results so far, to produce the human detection sets for all of our test cases and those used in producing comparison results. Furthermore, using the same method to produce the input set (ACF) makes the comparison between methods meaningful. Our system is an online tracking, which is quite different from batch tracking systems that can use future frames to correct past frames and thus gain higher tracking accuracy. Online multi-target tracking systems allow use of all information up to the current frame to predict target's state in the next frames. These systems could use various tracking components/cues such as: motion, shape and appearances combined in the same tracking system.

A two-stage association tracking framework described in [7] classifies targets into two types: a high confidence and a low confidence. It calculates a confidence level for each target by using the length of target's trajectory and average similarity score between a target and associated detections. A target is called "high confident" (i.e. highly reliable) if its confidence level is greater than a threshold. Otherwise, it is low confident. High confident targets are assigned as human detections in the first association stage while the low confident targets are further compared with all remaining targets and detections in the second stage to determine their probable associations. We implement several improvements to the two-stage target confident appearance framework [7] to implement our new spatiotemporal STKSVD dictionary learning method for online target's appearance learning and show results that are more accurate than the existing methods. It is important to note that the proposed STKSVD is framework-free and can easily be used with other tracking frameworks or combined with different tracking components.

**Discriminative Dictionary Learning:** Most optimization algorithms for dictionary learning to solve classification/recognition problems focus on face recognition applications [19], [20]. In multi-target tracking domain, sparse representation (i.e. using accelerated proximal gradient descent method [21])of each detection is mainly used to find residual error between a detection and targets as [22], [23]. These residual errors are directly used for calculating appearance similarity score. These methods collect all or some target samples from previous frame to construct a dictionary for each target but none have employed spatiotemporal appearance learning steps. A Label Consistent KSVD (LCKSVD) method proposed in [8] attempted to apply a face recognition appearance learning technique directly to multi-target tracking problem. They did not incorporate the key differences between the two problems and did not produce satisfactory results in terms of accuracy.

## III. SYSTEM DESIGN

### A. System Overview

A high level description of our system design and tracking algorithm are shown in Figure 1 and Algorithm 1, respectively. We define the following terms used throughout the paper: (1) Detection, the output of the "human detector" (ACF) algorithm, consists of the bounding box and its encompassing human object plus size, time, and location. (2) Target, a detected human that has been learned and entered in the dictionary. (3) Samples, entries in the dictionary corresponding to the same target, which can be due to different detections (variations). Furthermore, throughout the paper we refer to one column of the dictionary regardless of whether it corresponds to different targets or samples of the same target as an "atom".

**Initialization**: The dictionary is initialized using the first few frames (#initFrames = 5 in our test cases) to generate a number of targets. A target is generated by comparing the overlap between detected humans in frame #1 with subsequent frames (1&2, 2&3, 3&4, 4&5). When overlap is within a threshold (0.5) the detection is entered in the dictionary as a target. These initial targets are assigned a 0.75 value for "high confident" targets. Once we have the initial targets, we apply our STKSVD algorithm to learn appearances of these initial targets.

**Overview of Multi-target Tracking Steps:** For each frame in the input video sequence, we first apply ACF human detector [24] to detect all people in the scene. In the feature extraction step, we scale all human detection bounding boxes to the same scale (32x64) and calculate color histograms on the upper and lower body part (i.e. the top and bottom half of the bounding box) for each detection. We also extract size (before scaling) and location features of each detection in order to calculate the shape similarity $S_s$ and location similarity $S_l$ with existing dictionary targets in the association stages. Given color histogram feature of each human detection and learned dictionary D of all existing targets, we calculate the sparse representation of each detection using Orthogonal Matching Pursuit (OMP) algorithm [25]. OMP algorithm computes the nonlinear approximation (i.e. sparse representation) of a signal (e.g. appearance feature in our case) in a dictionary D. Sparse representation brings up the salient features that help in discriminating among different targets. Note that our dictionary consists of learned atoms of all targets from high confidence targets (red columns) to low confidence targets (blue columns) as shown in figure 1.

In the first stage of association, sparse representation of each detection is used as an input to the linear classifier in order to produce appearance similarity score between a detection and all targets. We multiply location, shape and appearance similarities to get the final similarity score. Then, the inverse of final similarity score is used as the cost (i.e. an element of the cost matrix) between a target and a detection. The dimensions of the cost matrix in this stage are $f \times n$, where $f$ is the number of targets and $n$ is the number of detections. Next, we apply the Hungarian algorithm [12] to the cost matrix to find the matched pairs between detections and targets. The Hungarian algorithm generally solves an assignment problem by finding pairs (pairs of targets and detection in our case) that minimize the total cost.

In the second stage of association, the low confident targets are assigned to either the remaining detections or one of the high confident targets. In this stage, we again use location, shape and appearance similarity to construct a secondary cost matrix. The dimensions of this cost matrix are the number of low confident targets ($f_l$) by the number of remaining detections plus the number of high confident targets ($f_h$). The difference in computation in this stage is that the inverse of the minimum residual reconstruction error is used as appearance similarity score instead of using linear classifier function. After the two stages of associations, we update the existing targets' states which include velocity, location and confidence level. Finally, our STKSVD algorithm is applied to learn appearance of existing targets, which results in updated learned dictionary $D$ and linear classifier parameters $W$.

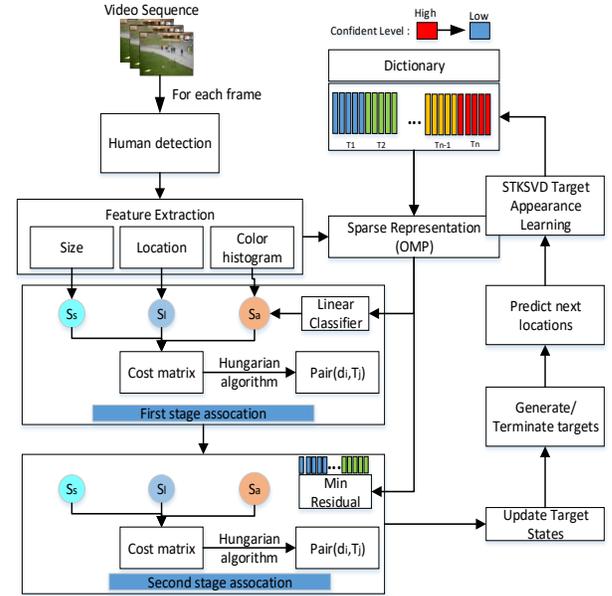

Figure 1. System Design

Next, we will present our calculations of confidence level, similarity score and target generation/termination.

**Target Confidence:** The idea is to use the target's computed confidence to set a higher priority for "high confident" targets to associate with human detection set in first stage. Then the remaining human detections are matched with low confident targets in the second stage. We define parameters amplifier $\alpha$, and attenuate $\beta$, to promote and demote the affinity score when there is a match between a detection and target. In in our experiments, we set a fixed value for $\alpha = 1.5$ and $\beta = 0.8$. $\alpha$ increases the target's confidence level when there is re-identification of the target while $\beta$ decreases the confidence level of a target when it is not detected. We reset the confidence level of a low-confident target to its initial value (0.75 in our experiments) in the current frame upon a new detection of this target. We observe that it helps recover the state of this target faster after occlusions. In summary, the confidence score of target $\tau_i$ at current frame if there is association between a detection $d_j$ and $\tau_i$ is updated as:

$$\text{Conf}(\tau_i) = \frac{1}{L}\Sigma_{t=t_s}^{t=t_c} S(\tau_i, d_j) * (1 - \exp(\sqrt{L}) \quad (1)$$

Where $t_s$ and $t_c$ are start frame and current frame, respectively. L is the length of target's trajectory $L = t_c - t_s$. $S(\tau_i, d_j)$ is similarity score between detection $d_j$ and target $\tau_i$, which is amplified or attenuated using $\alpha$ or $\beta$.

```
Algorithm 1: Multi-target tracking algorithm
 Input:
 • Learned dictionary $D = \{a_1, ..., a_K\} \in R^{N \times K}$ of all targets.
 • Learned linear classifier parameters $W$
 • Existing target $\tau = \{\tau_1, ..., \tau_f\}$, where f is the number of
   targets.
 •       A set of human detection $P = \{d_1, ..., d_n\}$, where n
   is the number of detections.
 Output:
 • Continuously track all targets
 • Generate new targets.

 1. For frame = #initFrames + 1 to lastFrame
 2. Extract color histogram feature $F_i \in R^{N \times 1}$ for each human
    detection
 3. Find matched detections for high confident target.
    (first association stage)
 4. On remaining detections, find matched ones for low
    confident targets. (second association stage)
 5. Update confidence value for all targets.
 6. Update target state  and predict location in the next frame
 7. Generate/ terminate targets
 8. STKSVD target appearance learning.
 9. End For
```

**Similarity Score**: For each stage of target-detection association, we define a similarity score between detection and target using shape similarity $S_s$, position similarity $S_p$, and appearance similarity $S_a$. The overall similarity score is $S = S_s S_a S_p$. The predicted position, $p_{tc} = p_{tc-1} + v_{tc-1} \times t$, of a target from last frame is used to estimate position similarity using a Gaussian distribution as: $S_p = \frac{1}{\sqrt{2\pi\sigma^2}} e^{-\frac{(p_{tc}-p_{dc})^{\wedge}2}{2\sigma^2}}$, where $p_{dc}$ is the position of a detection in the current frame, $v_{tc-1}$ is velocity of the target in previous frame and $\sigma = [\sigma_x, \sigma_y]$ is the spatial variance in 2D coordinates ($\sigma_x = 75, \sigma_y = 50$ in our experiments) . For shape similarity, we use $S_s(X, Y) = \exp(-\{\frac{h_X - h_Y}{h_X + h_Y} + \frac{w_X - w_Y}{w_X + w_Y}\})$ as in [7]. Where X and Y are target or detection, h and w indicate height and width of the detection or target.

Given the learned set of parameters $W = \{\omega_1, ..., \omega_n\} \in R^{n \times K}$ and dictionary D, we calculate the appearance similarity score $S_a$ between a target $\tau_i$ and detection $d_j$ for each of the two association stages.

In the first stage, we calculate the sparse codes $x_j$ for each detection's appearance features using OMP algorithm [25]. Then, the sparse codes are used as feature inputs in a linear classifier to find appearance score similarity between each detection and the existing high confident targets as:

$$S_a(\tau_i, d_j) = Wx_j \quad 1 \le i \le f \, \& \, 1 \le j \le n \quad (2)$$

Where f is the number of targets and n is the number of detections.

In the second stage, we calculate a similarity score to assign a low confident target to the remaining detections or other targets in the dictionary. Thus, only the part of the dictionary consisting of atoms of low-confident targets are needed for this calculation. We use the inverse of minimum reconstruction error for appearance similarity score as it will give the maximum probability of similarity for this set of targets and detections. Given a dictionary of low confident targets $D_L \subset D$, sparse codes $x_j$ of each detection's features $y_j^d$ or existing target features $y_j^{\tau e}$ are calculated as in stage one. Then, the residual error of each detection/existing target $y_j = \{y_j^d, y_j^{\tau e}\}$ on different low confident targets $\tau_i^l$ is calculated as:

$$r_{ij} = \|y_j - D_L x_{ij}\|_2^2 \quad (3)$$

where $x_{ij}$ is part of $x_j$ for a specific target class by keeping the coefficients for all dictionary atoms in target class $\tau_i^l$, and setting 0 for other classes. The appearance similarity between detection $d_j$ /existing target $\tau_j^e$ and target $\tau_i^l$ is defined as: $S_a(\tau_i^l, d_j/\tau_j^e) = \frac{1}{r_{ij}}$.

*B. Generation/Termination of Target*

**Generation**: Since we cannot make any assumptions about information related to exits/entries of targets in a scene, it is important to have a reliable target generation method in our multi-target tracking system.  Most existing methods assign a new target for any unassigned detection in a frame. This increases the chance of creating new false targets because of the imperfection of human detection sets (e.g. a target may be covered by several overlapping bounding boxes). Bae et al., in [7] generate a new target by considering the past predefined number of frames and look for the continuous overlap and height similarity in these frames. We generate a new target by considering its overlapped trajectory with existing target trajectories to determine whether the ratio of overlap between the pairs of trajectories is less than a pre-defined threshold.

**Termination**: We terminate tracking a target if it has not been detected in a fixed consecutive number of frames (5 in this implementation). This usually occurs when a target is under occlusion for a period of time greater than a predefined number of frames or if it is out of the scene. The re-identification of a target when it is out of the scene is in general not applied within the multi-target tracking research domain.

IV. ONLINE APPEARANCE LEARNING USING SPATIOTEMPORAL STKSVD DICTIONARY

Dictionary learning may be divided in two categories of (a) reconstruction where one attempts to learn the dictionary by finding sparse representation of input signals that minimize the reconstruction error[13], [26] and (b) classification where the dictionary learning combines the reconstruction error minimization simultaneously with a linear classifier [8], [14].

Our proposed STKSVD algorithm learns dictionary D of all targets and linear classifier parameters $W$ to reduce reconstruction and classification errors. We develop the STKSVD algorithm by incorporating spatiotemporal features

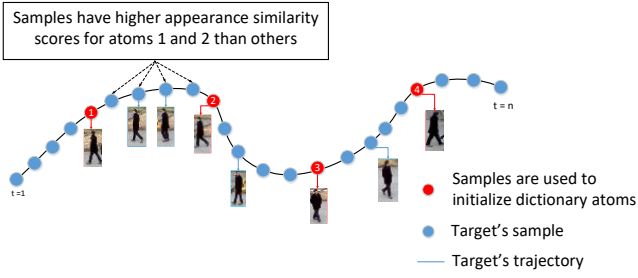

Figure 2. An example showing how atoms for a target $\tau$ are chosen to update the dictionary. It also illustrate the appearance similarity between an atom and its nearby samples. Each circle represents a detected sample in a frame for $\tau$.

so that it learns dictionary D and linear classifier parameters W simultaneously as formulated below:

$$< D, X, W > = \underset{D,X,W}{\mathrm{argmin}} \|Y - DX\|_2^2 + \kappa \|Q - AX\|_2^2 \\ + \lambda \|H - WX\|_2^2 \quad s.t. \; \forall i, \|x_i\|_0 \leq T \quad (4)$$

where $\|Y - DX\|_2^2$ is the reconstruction error term. $\|H - WX\|_2^2$ is the classification error. $\kappa$ and $\lambda$ are scalars that control the contribution of each term. The columns of the label matrix $H = \{h_1, \ldots, h_n\}$ indicate which class each of the input signals belong to. For example, $h_1 = \{1, 0, \ldots 0\}$ indicates that input signal $y_1$ belongs to class 1. $\|Q - AX\|_2^2$ is the sparse code error term, where we enforce sparse code to be discriminative between targets with respect to spatial and temporal features. A is a linear transformation matrix transforming the original sparse codes to the more discriminative sparse codes. Different from existing work, we construct discriminative sparse code matrix Q from two matrices $Q_l$ and $Q_{st}$ satisfying each of the following two criteria: (1) Each sparse code should have label consistency, which means it should have strong bias to the atoms in the same class. We expect that each of the input signals can be constructed by atoms in same class. We denote this sparse code matrix as $Q_l = \{q_1, \ldots, q_n\}$ where column $q_i$ is the expected sparse code of signal $y_i$. For example, in column $q_i = \{0, \ldots, 1, 1, \ldots 0\}$ a 0 in position k indicates that signal $y_i$ and the dictionary atom $d_k$ in D do not belong to the same class. (2) Associating a detected bounding box with a target should take into account the time and location of the detected boxes in order to match it with the best target and learn more effectively when several similarities are possible. To account for this spatiotemporal improvement, we denote $Q_{st} = \{\theta_{ij}\}, i = \{1 \ldots K\}, j = \{1 \ldots N\}$ spatial and temporal similarity score matrix between atoms and training signals, where each of element $\theta_{ij}$ is defined as:

$$\theta_{ij} = e^{(-\frac{|t_i - t_j| * \|p_i - p_j\|_2}{\sigma_s})} \quad (5)$$

where t and p are the time and position of the target's training sample and atom. $\sigma_s$ is the variance of spatial information. Finally, our discriminative sparse code Q can be calculated as element-wise multiplication of two matrices $Q_l$ and $Q_{st}$:

$$Q = Q_l \otimes Q_{st} \quad (6)$$

**Optimization**: Equation (4) can be approximated using the original KSVD algorithm as:

$$< D', X > = \underset{D,X,W}{\mathrm{argmin}} \|Y' - D'X\|_2^2 \quad s.t \; \forall i, \|x_i\|_0 \leq T \quad (7)$$

where $Y' = (Y^T, \sqrt{\kappa}Q^T, \sqrt{\lambda}H^T)$ and $D' = (Y^T, \sqrt{\kappa}A^T, \sqrt{\lambda}W^T)$. [8],[14] provide more details of optimization algorithms used for KSVD. The training is done for each target by using all detected samples for a given target from all previous frames up to the current frame.

For a given target, we select the detected sample in the trajectory of that target (temporal and spatial) that best represent it within the last few frames (five in our case) plus within all target samples already in the dictionary. This is the sample that is stored in the dictionary with the same class of the target. Figure 2, shows target samples (in red) that are stored in the dictionary as training continues for a specific target $\tau$. For example, atom 4 in the figure is selected from detected samples from the last five frames and atoms 1, 2, and 3 (already in the dictionary) representing samples of $\tau$. The linear classifier parameters W are initialized by solving the quadratic loss and $L_2$ norm regularization function as described in [27]:

$$W = \underset{W}{\mathrm{argmin}} \|H - WX\|_2^2 + \xi \|W\|_2^2 \quad (8)$$

By using regression model, Equation 8 is solved as: $W = (XX^T + \xi I)^{-1} XH^T$. A similar technique is applied for initializing A as: $A = (XX^T + \xi I)^{-1} XQ^T$.

## V. EXPERIMENTS

**Implementation Details**: We have implemented our system on a Xeon CPU E5-2650 v4@2.2GHz and utilized OMP and KSVD toolbox [28]. In this section, we present the results of three experiments: (a) Explore the impact of appearance learning model on overall tracking systems by varying the number of atoms. (b) Compare four existing appearance learning methods with our STKSVD. (c) Compare the overall accuracy of two existing target appearance learning-based tracking systems with our system. For meaningful comparison, all systems being tested use 2DMOT2015 dataset [29], which provides the same human detection set (i.e. resulted from ACF detector [24]) and evaluation tools.

For all experiments, we set the following values for the system parameters as follows: The association scores in both stages is set to equal 0.4. Target samples are scaled to size (64, 32) and their cascaded RGB color histogram is set to 48 bins so that variation of target sizes do not impact target appearance variations. In STKSVD learning process, we also set spatiotemporal constraint term $\kappa = 2$ and classification error term $\lambda = 4$.

**Datasets and Detections:** We use] 2DMOT2015 dataset for performance evaluation. Depending on the ground truth of each sequence and published results available, we set up two different experiments. The first experiment is conducted on sequences that have made their ground truth available so that we can observe the results on different parameters and compare different learning appearance techniques in our tracking system. The name of sequences used in this experiment are in Figure 3. In the second experiment, we compare our results with related previous work so we choose a set of sequences for which implementation and results are available. The names of these sequences are indicated in each experiment. In general, all selected sequences cover all possible scenarios that may occur in daily life such as static camera sequences (e.g. PETS09-S2L1, PETS09-S2L2, KITTI-16), moving camera sequence (e.g. ETH sequences), sparse crowd (e.g. PETS09-S2L1), medium crowd (e.g. ETHBahnof, KITTI-16) dense crowd (e.g. PETS09-S2L2), and also from different angles and positions of camera views. These sequences cover most possible challenges we may encounter in multi-target tracking. Since the performance of current tracking-by detection method depends significantly on the performance of human detectors, we believe that tracking algorithms should use the same human detector. Thus, we use the same publicly available detection method (ACF detector) and compare only the tracking systems.

**Evaluation Metrics**: We use the CLEAR MOT [30], which consists of multiple metrics for evaluations as follows:(1) MOTP (multiple object tracking precision) evaluates the intersection over the union area between a detected bounding box and the ground truth bounding box. (2) MOTA (multiple object tracking accuracy) is calculated using three sources of errors: false negative (FN), false positive (FP), and ID switch (IDS). (3) FAF is the average false alarms or false positive (FP) per frame. (4) MT (mostly tracked targets) is the ratio of ground-truth trajectories (GTT) covered by track hypothesis for at least 80% of their life span. (5) ML (mostly lost) is the ratio of ground-truth trajectories (GTT) covered by track hypothesis for at most 20% of their life span. (6) FP is the total number of false positives in a sequence. (7) FN is the total number of false negatives in a sequence. (8) IDS is the total number of identity switches. (9) Frag is the total number of times a trajectory is fragmented (i.e. interrupted during tracking).

**Analysis on different number of atoms:** In this experiment, we vary the number of atoms and observe the impact of it on the accuracy of our system. The purpose of this experiment is to find the number of atoms for which STKSVD generates most accurate results for each sequence. We use the resulting number of atoms in the next two sets of experiments (circled on Figure 3). We set the STKSVD appearance learning iterations to 10 and vary the number of atoms for each target in the dictionary from 5 to 40 and observe the results on 11 video sequences as listed in Figure 3. Note that our work focuses on appearance learning but only linear motion predictions. We observe that our system achieves better results in sequences that have the following features: (1) consisting of high number of targets that move linearly, (2) targets that have long trajectories so that our system can collect enough training data. Our best results are on PETS09-S2L1 where most of the targets have long trajectories with linear movements and fewer occlusions than other sequences. In addition, the higher number of atoms positively improves the accuracy in this sequence. If the number of atoms is too low, there won't be enough samples to represent the salient features of the target, i.e., MOTA is low (70%). When the number of atoms is too large, the variance of target appearances becomes too large which in turn reduces the accuracy. For example, when the number of atoms are 40, tracking accuracy is 74.5% while we obtain a higher accuracy of 78.2% when the number of atoms are 30. We do not observe a similar impact from varying the number of atoms used on the rest of sequences. This is because the trajectories of targets in these sequences are often short. We further observe that our system does not perform well in sequences that appearances of targets are difficult to distinguish such as when targets are usually small, affected by dark light, or are blurred by abrupt camera motions. The tracking accuracy of these sequences (ADL-Rundle-8 and Venice-2) are -6 % and -11 % respectively. Note that negative results mean that the sum of FP, FN and IDS is larger than the ground truth detections, which is mostly due to large FP and FN.

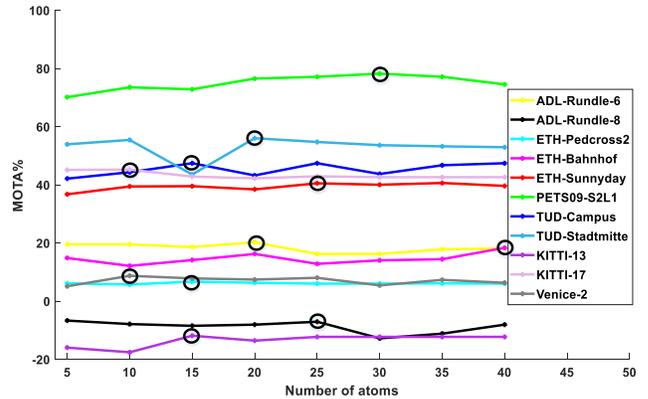

Figure 3. Effects of different number of dictionary oms on tracking accuracy.

**Comparison with four appearance learning algorithms**

In this section, we compare the results of our appearance learning model, STKSVD, with several known appearance learning algorithms: LCKSVD [8], ILDA [7], KSVD [13], and [7] which uses the Bhattacharyya distance using the set of sequences in Figure 3. All the above algorithms are run within the same two-stage association tracking system.

Table 1 shows the higher average accuracy of using our proposed STKVD on the 11 video sequences. The STKSVD achieves 20% accuracy while Bhattacharyya distance, LCKSVD, and ILDA achieve 18.4%, 17.5 % and 17.5 % accuracy, respectively. It is also of interest that STKVD also has better average ID switch metric for the video sequences. Figure 4 compares several scenarios where STKSVD produces more accurate results than its counterpart. In Figure 4-a, STKSVD can adaptively re-identify the missing target "10" under partial occlusion while LCKSVD results in a false detection (on target "6") and misses target "10" altogether. Figure 4-b shows an example of when STKSVD perform better during camera's abrupt motion in sequence

| Learning methods | MOTA%↑ | MOTP%↑ | FAF↓ | MT%↑ | ML%↓ | FP↓ | FN↓ | IDS↓ | Frag↓ |
|---|---|---|---|---|---|---|---|---|---|
| Bhattacharyya distance | 18.4 | 71.6 | 2.13 | 17.40 | 45.60 | 11718 | 20686 | 300 | 687 |
| KSVD | 17.9 | 71.4 | 2.12 | 17.6 | 46.4 | 11652 | 20917 | 316 | 698 |
| ILDA | 17.5 | **71.7** | 2.16 | 16.60 | 46.60 | 11854 | 20883 | 314 | 687 |
| LCKSVD | 17.5 | 71.3 | 2.27 | **25.2** | **44.8** | 12500 | 20253 | 295 | 767 |
| STKSVD | **20.0** | 71.5 | **2.12** | 17 | 45.4 | **11652** | **20151** | **261** | **750** |

TABLE 1. COMPARISON BETWEEN DIFFERENT DICTIONARY LEARNING ALGORITHMS ON TRAINING SEQUENCES (11 VIDEOS)

ETH-SunnyDay (a sequence recorded with a mobile device). As we see, STKSVD is able to re-identify a small target "9" while LCKSVD completely misses it. Figure 4-c shows a case where STKSVD outperforms ILDA when target "10" changes posture.

**Tracking System Comparisons:** In this section, we compare STKSVD with other tracking systems: TC_ODAL [4] and GSCR [12]. TC_ODAL is a tracking system that is not a dictionary learning-based method but uses target confidence and incremental linear discriminant analysis, ILDA, for online target appearance learning. GSCR uses accelerated proximal gradient algorithm for learning dictionary D. Similar to our work, the tracking accuracy of these methods are mostly impacted by a target appearance learning model. They also use Kalman filter for linear motion prediction. The experiment is conducted on several sequences in 2DMOT2015 dataset-as shown in Table 2 and shows that STKSVD achieves higher accuracy than other methods in every case. Furthermore, we observe better (i.e. lower) FNs and IDS in several sequences with STKSVD except for the PETS09S2L1 sequence. However, in PETS09S2L1 we achieve lower FPs, which also results in a higher accuracy in general.

## VI. DISCUSSION & CONCLUSION

We have developed a new spatial-temporal KSVD dictionary learning algorithm STKSVD, for target appearance learning to solve online multi-target problem. STKSVD characterizes the spatial and temporal dependencies between training data and dictionary atoms. Thus, it better learns a discriminative dictionary. We also improve the overall appearance learning method by using two different methods to calculate the appearance score similarity between detections and targets in a two-stage associations system. Our experiments show that STKSVD outperforms its counterpart appearance learning methods and tracking systems. There remain several limitations to be addressed as we further development our work:

- The performance of tracking-by-detection methods are significantly affected by human detector performance. ACF human detector that we used in this work often fails to detect up-close and large targets, which leads to tracking failures. It is reported that ACF detectors have on average about 60% precision and 46% recall on the sequences tested [29]. We believe that as better human detection or development of a joint tracking and detection frameworks will improve the tracking performance significantly.

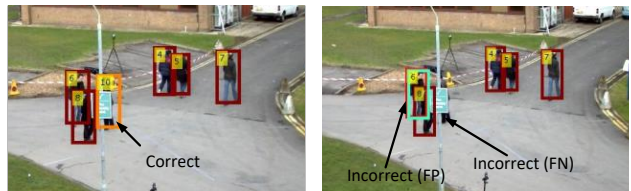

(a) Examples on PETS09S2L1 frame 12, where STKVD (left) correctly tracks under short-term occlusion compared to LCKSVD (right)

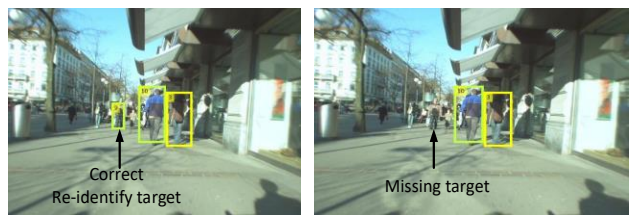

(b) Examples on ETH-Sunnyday frame 82, where STKVD (left) keeps tracking a small target correctly, while LCKSVD (right) lost this target

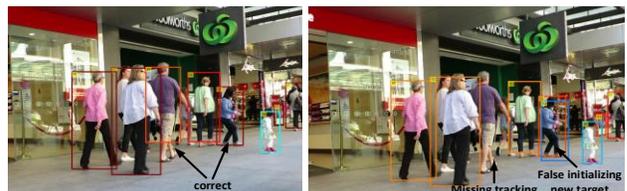

(c) In ADL-Rundle 6 frame 278, STKVD (left) shows better performance compared to ILDA (right) when it is able to track target in partial occlusion and another target while changing posture

Figure 4. Some example results of STKVD compared to other learning methods. Frames are slightly cropped out for better viewing

- In this work we predict human detections linearly using Kalman filtering [31] which often fails under long-term and multi-target occlusions. These are often scenarios where people's movements are unpredictable. Using STKSVD with non-linear motion models (e.g., [4], [5]) will result in improved tracking performance.

| Dataset | Methods | MOTA%↑ | MOTP%↑ | FAF↓ | MT%↑ | ML%↓ | FP↓ | FN↓ | IDS↓ | Frag↓ |
|---|---|---|---|---|---|---|---|---|---|---|
| PETS09S2L1 | TC ODAL [4] | 69.5 | 70.7 | 1.12 | **89.5** | 10.5 | 890 | 483 | 45 | **112** |
| | GSCR [12] | 69.9 | **71.2** | - | - | - | 805 | **557** | 35 | - |
| | STKSVD | **78.2** | 71 | **0.61** | 84.2 | 15.8 | **598** | 682 | **25** | 120 |
| PETS09S2L2 | TC ODAL [4] | 30.2 | 69.2 | 2.5 | 2.4 | 19 | 1074 | 5375 | 284 | 499 |
| | GSCR [12] | 24.1 | 67.6 | **2.2** | 0 | 28.6 | **946** | 6214 | **162** | **275** |
| | STKSVD | 31.6 | 68.2 | 3.0 | **4.8** | 14.3 | 1318 | **5099** | 176 | 339 |
| ETHMS (Jelmoli, Linthescher, Crossing) | TC ODAL [4] | 20.5 | **72.9** | 0.5 | 4.7 | 61.7 | 285.3 | 3116 | 17 | 56 |
| | GSCR [12] | 11.4 | 70.5 | **0.3** | 0.9 | 72.5 | 185.7 | 3582 | 14 | **37** |
| | STKSVD | **20.9** | 72.1 | 0.8 | **11.0** | 55.3 | 415 | **2973** | 13 | 46.7 |
| KITTI-16 | TC ODAL [4] | 27.3 | 70.7 | 1.0 | 0.0 | 17.6 | 213 | 986 | 37 | 106 |
| | GSCR [12] | 23.8 | 70.9 | **0.5** | 0.0 | 35.3 | **96** | 1,185 | 16 | **31** |
| | STKSVD | **28.3** | **72.7** | 1.4 | 0.0 | 11.8 | 298 | **907** | 14 | 63 |
| TUD-Crossing | TC ODAL [4] | 55.8 | 72.8 | 0.5 | 23.1 | 7.7 | 110 | 360 | 17 | 35 |
| | GSCR [12] | 35.2 | 72.7 | **0.3** | 0.0 | 23.1 | **70** | 628 | 16 | 22 |
| | STKSVD | **59.3** | **73.3** | 0.5 | **38.5** | 7.7 | 105 | **334** | **10** | **21** |

TABLE 2. COMPARISONS OF DIFFERENT RELATED TRACKING SYSTEMS ON CLEAR MOT METRICS